\PassOptionsToPackage{table}{xcolor}
\documentclass{article} 
\usepackage{iclr2025_conference,times}

\usepackage{amsmath,amsfonts,bm}

\def\eqref#1{equation~\ref{#1}}

\def\1{\bm{1}}

\DeclareMathAlphabet{\mathsfit}{\encodingdefault}{\sfdefault}{m}{sl}
\SetMathAlphabet{\mathsfit}{bold}{\encodingdefault}{\sfdefault}{bx}{n}

\usepackage{hyperref}
\usepackage{url}
\usepackage{booktabs}
\usepackage{bbding}
\usepackage{wrapfig}
\usepackage{graphicx}
\usepackage{multirow}
\usepackage{algorithm}
\usepackage{algorithmic}
\usepackage[table]{xcolor}

\makeatletter
\newcommand\figcaption{\def\@captype{figure}\caption}
\newcommand\tabcaption{\def\@captype{table}\caption}
\makeatother

\title{Customize Your Visual Autoregressive Recipe with Set Autoregressive Modeling}

\author{Wenze Liu\textsuperscript{1,2}, Le Zhuo\textsuperscript{2}, Yi Xin\textsuperscript{2,3}, Sheng Xia\textsuperscript{3}, Peng Gao\textsuperscript{2}, Xiangyu Yue\textsuperscript{1}\thanks{Corresponding author} \\
\textsuperscript{1}MMLab, The Chinese University of Hong Kong~~~\textsuperscript{2}Shanghai AI Laboratory~~~\textsuperscript{3}Nanjing University \\
\url{https://github.com/poppuppy/SAR}
}

\iclrfinalcopy 
\begin{document}

\maketitle

\vspace{-10pt}
\begin{figure}[ht]
    \setlength{\abovecaptionskip}{-6pt}
    \centering
    \includegraphics[width=\linewidth]{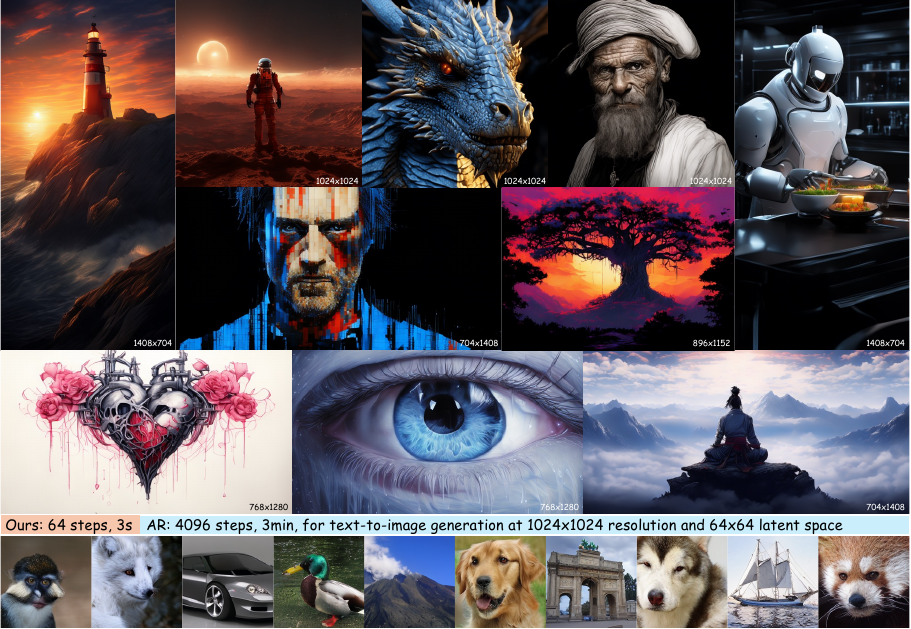}
    \caption{Text-conditioned and class-conditioned samples generated by SAR models. Our T2I model generates $1024\times1024$ images at a speed $60$ times faster than AR models.}
    \label{fig:visualization}
\end{figure}
\begin{abstract}
We introduce a new paradigm for AutoRegressive (AR) image generation, termed \emph{Set AutoRegressive Modeling} (SAR). SAR generalizes the conventional AR to the next-set setting, \textit{i.e.}, splitting the sequence into arbitrary sets containing multiple tokens, rather than outputting each token in a fixed raster order. To accommodate SAR, we develop a straightforward architecture termed \emph{Fully Masked Transformer}. We reveal that existing AR variants correspond to specific design choices of sequence order and output intervals within the SAR framework, with AR and Masked AR (MAR) as two extreme instances. Notably, SAR facilitates a seamless transition from AR to MAR, where intermediate states allow for training a causal model that benefits from both advantages of AR and MAR, such as few-step inference, KV cache acceleration and image editing. On the ImageNet benchmark, we carefully explore the properties of SAR by analyzing the impact of sequence order and output intervals on performance, as well as the generalization ability regarding inference order and steps. We further validate the potential of SAR by training a $900$M text-to-image model capable of synthesizing photo-realistic images with any resolution. We hope our work may inspire more exploration and application of AR-based modeling across diverse modalities. 
\end{abstract}

\section{Introduction}

The success of AutoRegressive (AR) models in Large Language Models (LLMs)~\citep{radford2018improving,radford2019language,brown2020language,raffel2020exploring,yang2019xlnet,touvron2023llama} has also driven their development in image generation, where some recent work~\citep{ramesh2021zero,yu2021vector,yu2022scaling,tian2024visual,li2024autoregressive,sun2024autoregressive,liu2024lumina} has demonstrated that the generative capabilities of AR models can rival or even surpass those of diffusion models~\citep{song2019generative,song2020score,ho2020denoising,dhariwal2021diffusion,rombach2022high,song2020denoising,lipman2022flow,liu2022flow,karras2022elucidating,peebles2023scalable,esser2024scaling,gao2024lumina,zhuo2024lumina}.

\begin{figure}[t]
    \setlength{\abovecaptionskip}{-3pt}
    \centering
    \includegraphics[width=\linewidth]{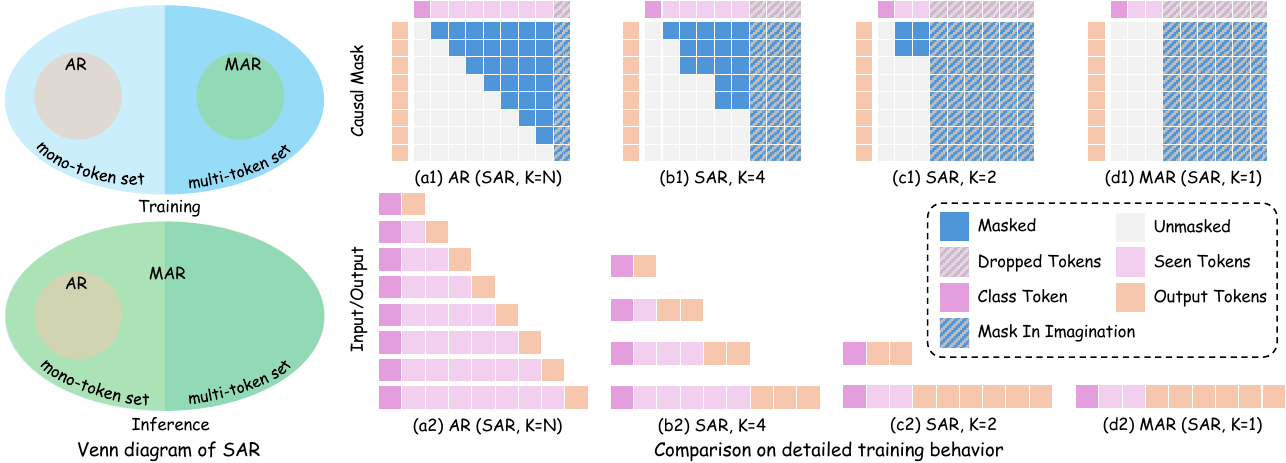}
    \caption{Conceptual illustration. SAR integrates existing AR variants by manipulating the sequence order and output intervals, creating a smooth transition path from classic AR to MAR.}
    \label{fig:comparison}
\end{figure}

Despite their strong performance, the large number of inference steps due to `next-token prediction' has become a bottleneck. This limitation has inspired some efficient approaches, with the idea of outputting multiple tokens simultaneously. Existing work~\citep{chang2022maskgit,yu2023magvit,chang2023muse,li2023mage,li2024autoregressive,ni2024revisiting} usually adopts BERT-like~\citep{devlin2018bert} masked modeling approaches to exchange the cost of always performing global computations (thus KV cache is not allowed) for fewer inference steps. Another stream of work designs proper sequence orders and arranges multiple tokens with similar properties into one group, to predict these tokens at once, \textit{e.g.}, the scale-aware order~\citep{tian2024visual,zhang2024var,ma2024star}. We conclude that, in the training phase, these approaches pay attention to two aspects: one is the \textit{sequence order}, the other is the \textit{output intervals}. The defined order and intervals split the sequence into token sets. AR splits the sequence into sets of single tokens, VAR~\citep{tian2024visual} builds multi-scale sets for an image, and Masked AR (MAR)~\citep{chang2022maskgit,li2023mage,li2024autoregressive} randomly divides the sequence into a masked set and an unmasked set. Fig.~\ref{fig:comparison} (a1, a2) illustrates examples for AR with intervals of length $1$, while (d1, d2) demonstrates MAR with $2$ output intervals.

In this work, we present \emph{Set AutoRegressive Modeling} (SAR), extending causal learning by generalizing sequence order and output intervals to arbitrary configurations. Specifically, compared with AR that splits the training process into sub-processes that output one single token in fixed raster order, SAR is able to input token sequence in any order (some examples are illustrated in Fig.~\ref{fig:rearrange} and Fig.~\ref{fig:orders}), and splits it into any number of token sets, each as a sub-process that output multiple tokens. In order to represent the sequential relationship of token sets, we introduce generalized causal masks. As shown in Fig.~\ref{fig:comparison}, the classical causal mask (a1) is a lower triangular matrix; when the set contains more than one token (b1, c1, d1), the matrix becomes block-wise and is called a generalized causal mask. Within our framework, we show that AR, VAR (analogously), and MAR emerge as special cases of SAR, with AR and MAR representing two extreme instances. Refer to the left side of Fig.~\ref{fig:comparison} and Table~\ref{tab:comparison} for conceptual illustrations. Moreover, with the new formulation, we offer a path for smoothly transiting between AR and MAR. The intermediate states of SAR enable one to train causal models that inherit both merits of AR and MAR, such as few-step inference, KV cache acceleration, and image editing.
Given that classical AR models, such as the decoder-only transformer, fail in the SAR setting, we propose a simple model architecture termed \emph{Fully Masked Transformer (FMT)}. FMT adopts the encoder-decoder structure proposed in the original transformer~\citep{vaswani2017attention} to enable both recording the output position and facilitating position-aware interaction between seen and output tokens. And it incorporates generalized causal masks into each attention process to keep the causal manner, and the details can be referred to Fig.~\ref{fig:model}. 

Under the SAR framework with FMT, we conduct experiments to explore the properties of SAR on the ImageNet $256\times256$ benchmark. We examine the effect of sequence order and output intervals on generation performance as well as the generalization ability across inference order and steps, and discuss the associated trade-offs. 
Then, we train a $900$M text-to-image model on $20$ million high-aesthetic images to further validate the generation potential of the transition states in SAR. Using limited computational resources and data, the trained model demonstrates the capability to produce photo-realistic images of arbitrary aspect ratios that adhere to the text descriptions. Its flexibility of outputting tokens in any order also enables effective application in zero-shot image editing tasks, such as inpainting and outpainting.

Our main contributions are:
\begin{enumerate}
    \item[i)] We propose Set AutoRegressive Modeling, that unifies existing AR variants and offers new states between the two extremes, AR and MAR. The new states introduce models that incorporate the advantages of both AR and MAR.
    \item[ii)] In line with SAR, we design a transformer model named Fully Masked Transformer, which enables causal learning with any sequence order and any output intervals.
    \item[iii)] We conduct extensive experiments to investigate the properties of SAR and the modeling capability of FMT. With a particular focus on the transition states, we explore the effectiveness of text-to-image generation and editing.
\end{enumerate}

\begin{table}[!t]
    \centering
    \begin{minipage}[ht]{0.35\textwidth}
    \setlength{\abovecaptionskip}{-5pt}
    \setlength{\belowcaptionskip}{-8pt}
        \centering
        \includegraphics[width=\linewidth]{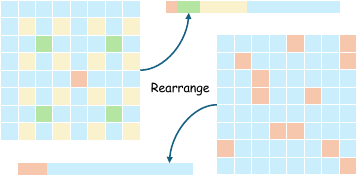}
        \figcaption{Sequence in any order can be rearranged as a causal one.}
        \label{fig:rearrange}
    \end{minipage}%
    \hfill
    \begin{minipage}[ht]{0.6\textwidth}
        \scriptsize
        \tabcaption{Comparison among existing autoregressive image generation paradigms. SAR is more flexible and enjoys merits of other paradigms.}
        \label{tab:comparison}
        \vspace{5pt}
        \centering
        \addtolength{\tabcolsep}{5pt}
        \begin{tabular}{@{}lcccc@{}}
            \toprule
            Method & AR & VAR & MAR & SAR\\
            \midrule
            Few-step inference & \XSolidBrush & \Checkmark & \Checkmark & \Checkmark \\
            KV cache & \Checkmark & \Checkmark & \XSolidBrush & \Checkmark \\ 
            Training/inference & Match & Match & Unmatch & Flexible \\
            Common VAE & \Checkmark & \XSolidBrush & \Checkmark & \Checkmark \\
            \bottomrule
        \end{tabular}
    \end{minipage}
\end{table}

\section{Related Work}
\subsection{Autoregressive and Masked Modeling}
Originating in language processing, GPT series~\citep{radford2018improving,radford2019language,brown2020language} and BERT~\citep{devlin2018bert} are representative works in autoregressive and masked modeling respectively. During the AR training, the current output token can only be observed by the preceding tokens. At inference tokens remain unchanged once output, facilitating the use of KV cache acceleration. Recently some work \citep{cai2024medusa,gloeckle2024better} studies to reduce the inference steps by training multiple prediction heads and conducting speculative decoding \citep{leviathan2023fast,chen2023accelerating} at inference. In contrast, BERT~\citep{devlin2018bert} employs a bidirectional modeling approach known as masked modeling, to capture contextual information. It randomly masks a portion of tokens at a high masking ratio and trains the model to predict these masked tokens. At inference, BERT models can iteratively generate the output sequence with fewer steps than AR methods, at the cost of global calculation. Additionally, some works have introduced context perception into AR models. For example, XLNet~\citep{yang2019xlnet} integrates insights from BERT by permuting the input sequence to enable bidirectional training with AR models. On image modality, our work not only provides further unification of AR and BERT models but also builds a smooth path connecting AR and BERT, where one can train models with both their merits. 

\subsection{Autoregressive Image Generation}
By tokenizing continuous images into discrete tokens using VQ-VAE~\citep{van2017neural,razavi2019generating,esser2021taming}, image synthesis can be accomplished by AR models~\citep{esser2021taming,lee2022autoregressive,ramesh2021zero,yu2021vector,yu2022scaling,liu2024lumina,luo2024open} just like language modeling. Recently, Llamagen~\citep{sun2024autoregressive} verifies the generation capability of plain LLM, Llama~\citep{touvron2023llama} on image modality. VAR~\citep{tian2024visual} divides the image latent space into several scale groups by training a multi-scale VAE, and conduct next-scale prediction. \citet{li2024autoregressive} point out that the BERT-like image generation models (\textit{e.g.}, MaskGIT~\citep{chang2022maskgit}, MagViT~\citep{yu2023magvit,yu2023language}, MAGE~\citep{li2023mage}, MAR~\citep{li2024autoregressive}) can also be regarded as autoregressive ones at inference, and as a result, we call BERT-like image generation models as MAR models. AutoNAT~\citep{ni2024revisiting} revisits and improves the designs of training and inference process of MAR models. \citet{li2024autoregressive} additionally show that autoregressive image generation can also be conducted on continuous latent space with diffusion loss. Our proposed SAR paradigm can encompass the existing approaches as special instances, and provide the users with more flexible design space regarding various trade-offs. The supporting model of SAR is built upon LlamaGen~\citep{sun2024autoregressive} for its plain nature.

\section{Method}
\label{sec:method}
In this section, we first review the AR and MAR paradigms. Then, we point out that conceptually these two methods differ in sequence order and output intervals, based on which we introduce Set AutoRegressive Modeling (SAR), and present the model design.
\subsection{Preliminary}
\textbf{AutoRegressive Modeling (AR).} AR models the distribution of a token sequence $\{x^1,x^2,...,x^n\}$ by the `next-token prediction' objective defined as
\begin{equation}
\label{eq:next_token}
    p(x^1,...,x^n)=\prod\limits_{i=1}^n p(x^i|x^1,...,x^{i-1})\,,
\end{equation}
where $p(x^1,x^2,...,x^n)$ is the probability density function.
Regarding the implementation, AR models are typically a decoder-only transformer with causal masks, as shown in Fig.~\ref{fig:comparison} (a1). During training, the input to the model is set as the sequence shifted by one position, \textit{i.e.}, dropping the last token, and padding a class token at the beginning (under the class-conditioned setting). The target is the original sequence, such that each output token is aligned with its `next token'. At inference, the model can output tokens one by one in an autoregressive manner. 

\begin{table}[t]\scriptsize
    \centering
    \begin{minipage}[ht]{0.47\textwidth}
        \tabcaption{Some examples on SAR setting.
        $\textrm{rand}(N,K)$ means randomly generate $K$ natural numbers, whose total sum is $N$. 
        }
    \vspace{1pt}
    \label{tab:hyperparams}
    \centering
    \renewcommand{\arraystretch}{1.1}
    \addtolength{\tabcolsep}{2.8pt}
    \begin{tabular}{@{}lcccccccc@{}}
    \toprule
        SAR & order & \#sets & intervals \\
        \midrule
        AR & raster & $N$ & $1,1,1,...$ \\
        VAR & custom & $log_4N+1$ & $1,4,16...$ \\
        MAR & random & 1 & $\textrm{rand}(N,2)$ \\
        Transition & random & K & $\textrm{rand}(N,K)$ \\
    \bottomrule
    \end{tabular}
    \end{minipage}%
    \hfill
    \begin{minipage}[ht]{0.5\textwidth}
        \tabcaption{Model setting of Fully Masked Transformer. The numbers of encoder and decoder layers are set equal for simplicity. Other configurations follows LlamaGen~\citep{sun2024autoregressive}.}
    \label{tab:model}
    \centering
    \renewcommand{\arraystretch}{1.1}
    \addtolength{\tabcolsep}{-1.5pt}
    \begin{tabular}{@{}lcccccccc@{}}
    \toprule
        SAR & Parameters & Enc. Layers & Dec. Layers & Width & Heads \\
        \midrule
        B & 125M & 6 & 6 & 768 & 12 \\
        L & 394M & 12 & 12 & 1024 & 16 \\
        XL & 893M & 18 & 18 & 1280 & 20 \\
    \bottomrule
    \end{tabular}
    \end{minipage}
\end{table}
\textbf{Masked AutoRegressive Modeling (MAR).} MAR has recently been abstracted by \citet{li2024autoregressive}, which describes the inference process of BERT-like~\citep{devlin2018bert} image generation methods (\citet{chang2022maskgit,li2023mage,yu2023magvit,yu2023language,li2024autoregressive}). In training, the input tokens are randomly masked with a high ratio (\textit{e.g.}, $70\%-100\%$ in \citet{li2024autoregressive}), and the model is trained to learn to predict the masked part. Fig.~\ref{fig:comparison} (a2) and (d2) illustrate that AR trains $n$ sub-processes in a single iteration, while MAR processes one sub-process at a time. At inference, these methods can predict multiple tokens at once, costing less number of steps than AR models. However, because the masked modeling process is not causal, it does not support causal techniques, \textit{e.g.}, KV cache acceleration. \citet{li2024autoregressive} define `next set-of-tokens prediction' as
\begin{equation}
\label{eq:next-set}
    p(x^1,...,x^n)=p(X^1,...,X^K)=\prod\limits_{k=1}^K p(X^k|X^1,...,X^{k-1})\,,
\end{equation}
where $X^k=\{x^i,x^{i+1},...,x^j\}$ is a \textit{set of tokens} to be predicted at the $k$-th step. Eq.~\ref{eq:next-set} generalizes vanilla next-token prediction Eq.~\ref{eq:next_token} at inference time. 

\subsection{Set AutoRegressive Modeling}

\textbf{Sequence order and output intervals characterize autoregressive paradigms.} Actually, the token sequence in any output order can be rearranged into a causal one. AR is the simplest case whose input sequence is inherently causal. The other two instances with respect to an $8\times8$ image token grid are shown in Fig.~\ref{fig:rearrange}. The left order is derived by downsampling the tokens using nearest neighbor interpolation (so the token values stay unchanged after interpolation). We make the model progressively output tokens downsampled with a scale factor of $1/8$, $1/4$, and $1/2$, and finally the rest of the tokens in a scale-aware order. It shares a similar spirit with VAR~\citep{tian2024visual}, so we call it a `next-scale' variant. In this case, we can rearrange the tokens in the scale order. The right subfigure corresponds to mask modeling. By putting the unmasked tokens at the front and masked ones as the rest, we also derive a causal sequence.

Next, we consider the output intervals. For example, the output intervals of the `next-scale' variant in Fig.~\ref{fig:rearrange} are $1,4,16,43$, while those of the masked variant are the number of masked tokens and unmasked tokens. Since these variants output multiple tokens in each interval, they should be paired with generalized causal masks in training. Some conceptual instances are shown in Fig.~\ref{fig:comparison} (b1, c1, d1), where generalized causal masks extend the classical causal mask (a1) to a block-wise format. The generalized causal mask can be uniquely determined by the output intervals.

\begin{figure}[t]
    \centering
    \begin{minipage}{0.5\textwidth}
        \centering
        \begin{algorithm}[H]
            \caption{SAR Training}
            \label{alg:sar_training}
            \begin{algorithmic}
                \STATE \textbf{Input:} Dataset $D$, Model $M$, Loss Function $\mathcal{L}$, Sequence Order \texttt{od}, Output Intervals \texttt{intv}
                \STATE \textbf{Output:} Model $M$
                \vspace{-10pt}
                \STATE \FOR{image code $x$, label $y$ \textbf{in} $D$}
                \STATE $x \leftarrow \textrm{rearrange}(x,\texttt{od})$, $t \leftarrow x$
                \STATE $x \leftarrow \textrm{drop\_last}(x,\texttt{intv}[-1])$
                \STATE $x \leftarrow \textrm{concat}(y,x)$
                \STATE $m_e, m_{ds}, m_{dc} \leftarrow \textrm{gen\_masks}(\texttt{intv})$
                \STATE $o \leftarrow M(x, m_e, m_{ds}, m_{dc},\texttt{od})$
                \STATE $l \leftarrow \mathcal{L}(o,t)$, backpropagate $l$
                \ENDFOR
                \RETURN $M$
            \end{algorithmic}
        \end{algorithm}
    \vspace{-8mm}
    \end{minipage}\hfill
    \begin{minipage}{0.45\textwidth}
        \centering
        \begin{algorithm}[H]
            \caption{SAR Inference}
            \label{alg:sar_inference}
            \begin{algorithmic}
                \STATE \textbf{Input:} Model $M$, Label $y$, Sequence Order \texttt{od}, Output Intervals \texttt{intv}
                \STATE \textbf{Output:} Image Code $x$
                \STATE $x \leftarrow \textrm{zero\_initialize}(\textrm{sum}(\texttt{intv}))$
                \STATE $m_e, m_{ds}, m_{dc} \leftarrow \textrm{gen\_masks}(\texttt{intv})$
                \vspace{-10pt}
                \STATE \FOR{$i$ \textbf{in} \texttt{intv}}
                    \STATE $o \leftarrow M(y,m_e, m_{ds}, m_{dc},\texttt{od},i)$
                    \STATE $z \leftarrow \textrm{sample}(o)$
                    \STATE $y \leftarrow \textrm{concat}(y,z)$
                    \STATE $x \leftarrow \textrm{scatter}(x,z,\texttt{od},i)$
                \ENDFOR
                \RETURN $x$
            \end{algorithmic}
        \end{algorithm}
    \vspace{-8mm}
    \end{minipage}
\end{figure}

SAR generalizes AR by extending the sequence order and the output intervals to any possible scenarios. In Fig.~\ref{fig:comparison} (a1, d1) we can see that the causal mask of AR and MAR are two extreme cases. In the intermediate states of SAR, one can train causal models with few-step inference enabled, which do not appear in either AR or MAR families. For example, if a $8$-token sequence is split into $4$ sets with $1,2,2,3$ tokens, the causal mask should be like that in Fig.~\ref{fig:comparison} (b1). In short, SAR extends `next-set prediction' in Eq.~\ref{eq:next-set} to the training phase.

\textbf{The model implementation---Fully Masked Transformer.} The realization of SAR is not straightforward, though. Classical AR models, \textit{e.g.}, the decoder-only transformer fail in three aspects. i) When AR shifts the sequence to align the current set with the previous set, it will find the number of tokens may not be equal. ii) AR models can only model the output-seen relations with fixed and simple `next token' forms of relative positional relationships, rendering them ineffective in complex scenarios involving arbitrary sets. iii) Given a token at a specific position, AR models output it based on its relative steps to the first token, leading to failure when outputting arbitrary sets. These drawbacks inspire the design philosophy: i) the model should have perception of absolute positions for outputting arbitrary token sets, and ii) the output tokens and the seen tokens should be placed into two containers, each with positional encoding, to facilitate their position-aware interaction. 

Starting from the AR model, we split the decoder-only transformer into two parts, an encoder and an decoder. The encoder takes in the image tokens and extract the semantic features. The decoder records the output position with position embeddings and models the interaction between output tokens and seen tokens from the encoder, at the cost of adding cross-attention in each decoder layer. 
\begin{wrapfigure}[28]{r}{0.5\linewidth}
    \setlength{\abovecaptionskip}{-4pt}
    \centering
    \includegraphics[width=\linewidth]{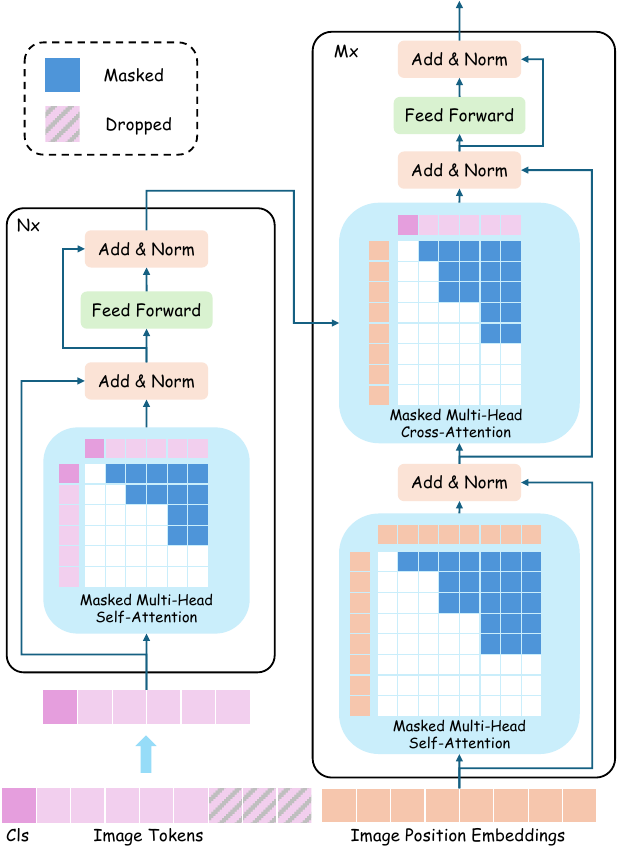}
    \caption{The model architecture of Fully Masked Transformer. Conceptually, it is the transformer in \citet{vaswani2017attention} plus generalized causal masks.}
    \label{fig:model}
\end{wrapfigure}
Additionally, generalized causal masks are added into each attention process, following the spirit of `the current token set to be predicted can only see the preceding sets'. In short, it can be regarded as a vanilla encoder-decoder transformer~\citep{vaswani2017attention} with generalized causal masks in all attention processes, as shown in Fig.~\ref{fig:model}. Consequently, we refer to it as the Fully Masked Transformer (FMT). Due to the fully causal feature, FMT naturally supports causal techniques like KV cache acceleration.

\textbf{The training procedure.} In order to train one model under the SAR framework, one should first specify the hyper-parameters, sequence order and output intervals. Based on the order setting, we first rearrange the sequence to the causal version (Fig.~\ref{fig:rearrange}). And we set the target as the rearranged causal sequence. Next, based on the output intervals, we drop the last set of the rearranged sequence and prepend a class token. The resulting sequence is then fed into the encoder. Then the model can be trained with the common cross entropy loss. We list several combinations of sequence order and output intervals in Table~\ref{tab:hyperparams}, where we also add the number of sets for better understanding. The overall training procedure is illustrated in Algorithm~\ref{alg:sar_training}.

\textbf{The inference configuration.} Since SAR is a generalized AR framework, it naturally supports advanced strategies developed for AR models, such as top-k, top-p, and min-p~\citep{nguyen2024min} sampling. In this work, we directly apply some simple strategies for inference; one may also customize their own inference schedules. The inference algorithm is summarized in Algorithm~\ref{alg:sar_inference}.

\section{Experiments}
\label{sec:experiments}

\subsection{Experimental Settings}
We conduct exploratory experiments on ImageNet~\citep{deng2009imagenet} $256\times256$ benchmark. We use the tokenizer provided by \citet{sun2024autoregressive}, and precompute the image codes before training as in \citet{sun2024autoregressive}. 
We always use a batch size of $256$ and learning rate of $1e-4$ during training. Models in the transition states SAR-TS in Table~\ref{tab:performance} is trained for $300$ epochs, while all other models are trained for $200$ epochs. Other training settings follow \citet{sun2024autoregressive}. For evaluation, we report the common used FID~\citep{heusel2017gans}, IS~\citep{salimans2016improved}, Precision and Recall metrics. Unless otherwise specified, the default setting is cfg=$2.0$, top-k=$0$ (all), top-p=$1.0$, temperature=$1.0$. The evaluation is conducted following \citet{dhariwal2021diffusion}. 

\subsection{Hyper-parameters in SAR}
\begin{wrapfigure}[9]{r}{0.6\linewidth}
    \setlength{\abovecaptionskip}{-5pt}
    \centering
    \includegraphics[width=\linewidth]{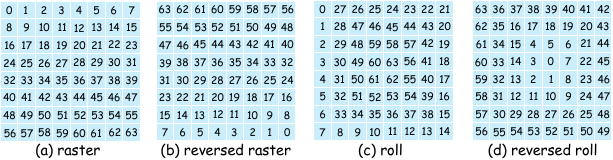}
    \caption{Some sequence order settings in the experiment. Taking the $8\times8$ case as illustration.}
    \label{fig:orders}
\end{wrapfigure}

\textbf{Configuration on sequence order and output intervals for training.} 
We test several hyper-parameter combinations containing some common settings and two customized ones named `next-scale' and `masked modeling'. Among the common settings, we control the sequence order, the output schedule, and the number of sets, where the latter two jointly determine the output intervals. There are six choices in order, among which the first four is shown in Fig.~\ref{fig:orders}. (a) The `raster' order is the classical AR order, while (b) is its reversed version. (c) and (d) are the `Swiss roll', clockwise, from outside to inside and from inside to outside respectively. The other two are fixed-random and random. The former means that we randomly generate an order and fix it during training, while the latter indicates generating random orders online at each training step. 
There are two types of output schedules involved, which can determine the output intervals based on the number of sets as follows: i) cosine: given a set number $K$, the output intervals $\{n_i\}_1^K$ follows $n_i=[N(\textrm{cos}(\frac{\pi}{2}\frac{i}{K})-\textrm{cos}(\frac{\pi}{2}\frac{i-1}{K}))]$, as in \citet{li2024autoregressive}. Note: here at least one token is ensured to be output at each step, thus given sequence order as raster and set number as $256$, it will recover to AR. ii) random: given a set number $K$, we randomly generate $K-1$ natural numbers (there may be equal numbers) between $0$ and $N$ with the same probability, such that the sequence can be split into $K$ intervals by these partition numbers. Under the common settings, we conduct experiments in the format of (sequence order)-(number of sets)-(output schedule). For example, raster-$64$-cosine indicates a raster-order sequence with $64$ sets under a cosine schedule.

The customized settings includes i) next-scale: we rearrange the $16\times16$ image tokens such that the $1$st set contains the $1/16$ nearest neighbor downsampled token, the $2$nd set contains the four $1/8$ downsampled tokens, ..., and the $5$th set contains the rest of tokens, as illustrated on the left of Fig.~\ref{fig:rearrange}, and ii) masked modeling: we follow the settings in \citet{li2024autoregressive}. Actually it can be derived by removing the loss of the first token set and modifying the random strategy in `random-$2$-random'.

\textbf{Configuration on model size of FMT.} The implementation of FMT is based on the GPT model in LlamaGen~\citep{sun2024autoregressive}. For simplicity, we do not adopt the asymmetric design in \citet{he2022masked}, but just divide the $N$-layer transformer into an encoder and a decoder, each with an equal number of layers. One can refer to Table~\ref{tab:model} for detailed model configurations.
Compared with LlamaGen, we add an extra cross-attention module at each decoder layer, so under the same model size, the number of parameters of FMT is slightly larger.

\vspace{-1mm}
\subsection{Main results}
\vspace{-1mm}
\begin{wrapfigure}[12]{r}{0.3\linewidth}
    \setlength{\abovecaptionskip}{-4pt}
    \centering
    \includegraphics[width=\linewidth]{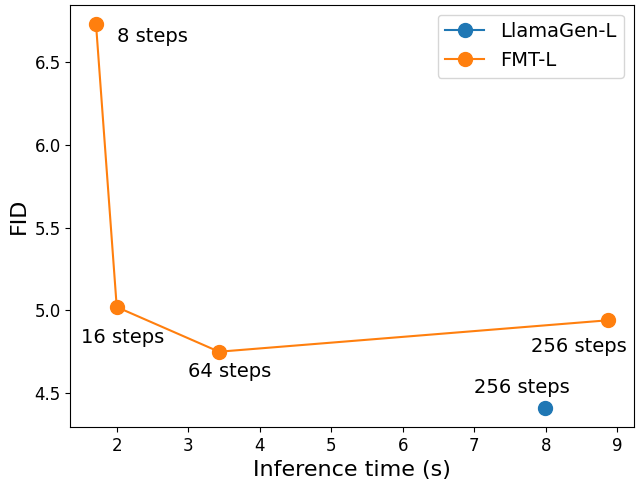}
    \caption{Trade-off between performance and time, using LlamaGen-L as a reference.}
    \label{fig:imagenet_time}
\end{wrapfigure}
\begin{table}[!t] \scriptsize
    \caption{Performance comparison among various paradigms and models. `-re' means rejection sampling. For LlamaGen~\citep{sun2024autoregressive}, * means direct training on $256\times256$ images; otherwise, training is on $384\times384$ and the output is resized in evaluation. `TS' denotes transition state.}
    \label{tab:performance}
    \centering
    \renewcommand{\arraystretch}{1.1}
    \addtolength{\tabcolsep}{4pt}
    \begin{tabular}{@{}lcccccccc@{}}
    \toprule
        Type & Model & \#Params & FID$\downarrow$ & IS$\uparrow$ & Precision$\uparrow$ & Recall$\uparrow$ \\
        \midrule
        \multirow{3}{*}{GAN} & BigGAN~\citep{brock2018large} & 112M & 6.95 & 224.5 & 0.89 & 0.38 \\
         & GigaGAN~\citep{kang2023scaling} & 569M & 3.45 & 225.5 & 0.84 & 0.61 \\
         & StyleGAN-XL~\citep{sauer2022stylegan} & 166M & 2.30 & 265.1 & 0.78 & 0.53 \\
        \midrule
        \multirow{4}{*}{Diffusion} & ADM~\citep{dhariwal2021diffusion} & 554M & 10.94 & 101.0 & 0.69 & 0.63 \\
         & CDM~\citep{ho2022cascaded} & - & 4.88 & 158.7 & - & - \\
         & LDM-4~\citep{rombach2022high} & 400M & 3.60 & 247.7 & - & - \\
         & DiT-XL/2~\citep{peebles2023scalable} & 675M & 2.27 & 278.2 & 0.83 & 0.57 \\
        \midrule
         \multirow{5}{*}{Masked AR} & MaskGIT~\citep{chang2022maskgit} & 227M & 6.18 & 182.1 & 0.80 & 0.51 \\ & MaskGIT-re~\citep{chang2022maskgit} & 227M & 4.02 & 355.6 & - & - \\
          & MAGE~\citep{li2023mage} & 230M & 6.93 & 195.8 & - & - \\
          & MAR-H~\citep{li2024autoregressive}  & 943M & 1.55 & 303.7 & 0.81 & 0.62 \\
         (SAR, K=1) &\cellcolor{gray!10}FMT-B  &\cellcolor{gray!10}125M &\cellcolor{gray!10}6.98 &\cellcolor{gray!10}222.28 &\cellcolor{gray!10}0.87 &\cellcolor{gray!10}0.36 \\
          &\cellcolor{gray!10}FMT-L &\cellcolor{gray!10}394M &\cellcolor{gray!10}6.13 &\cellcolor{gray!10}278.81 &\cellcolor{gray!10}0.88 &\cellcolor{gray!10}0.40 \\
        \midrule
         \multirow{3}{*}{VAR} & VAR-d30~\citep{tian2024visual} & 2.0B & 1.92 & 323.1 & 0.82 & 0.59 \\
          & VAR-d30-re~\citep{tian2024visual} & 2.0B & 1.80 & 356.4 & 0.83 & 0.57 \\
         (SAR, customized) &\cellcolor{gray!10}FMT-B &\cellcolor{gray!10}125M &\cellcolor{gray!10}12.49 &\cellcolor{gray!10}148.53 &\cellcolor{gray!10}0.76 &\cellcolor{gray!10}0.36 \\
        \midrule
        \multirow{14}{*}{AR} & VQGAN~\citep{esser2021taming} & 1.4B & 15.78 & 74.3 & - & - \\
        & VQGAN-re~\citep{esser2021taming} & 1.4B & 5.20 & 280.3 & - & - \\
         & ViT-VQGAN~\citep{yu2021vector} & 1.7B & 4.17 & 175.1 & - & - \\
         & ViT-VQGAN-re~\citep{yu2021vector} & 1.7B & 3.04 & 227.4 & - & - \\
         & RQTran.~\citep{lee2022autoregressive} & 3.8B & 7.55 & 134.0 & - & - \\
         & RQTran.-re~\citep{lee2022autoregressive} & 3.8B & 3.80 & 323.7 & - & - \\
         & LlamaGen-B* (cfg=2.00) & 111M & 5.46 & 193.61 & 0.84 & 0.46 \\
         & LlamaGen-L (cfg=2.00) & 343M & 3.07 & 256.06 & 0.83 & 0.52 \\
         & LlamaGen-XL (cfg=1.75) & 775M & 2.62 & 244.08 & 0.80 & 0.57 \\
         & LlamaGen-L* (cfg=2.00) & 343M & 4.41 & 288.17 & 0.86 & 0.48 \\
         & LlamaGen-XL* (cfg=1.75) & 775M & 3.39 & 227.08 & 0.81 & 0.54 \\
        (SAR, K=N) &\cellcolor{gray!10}FMT-B (cfg=2.00) &\cellcolor{gray!10}125M &\cellcolor{gray!10}5.40 &\cellcolor{gray!10}216.93 &\cellcolor{gray!10}0.87 &\cellcolor{gray!10}0.42 \\
         &\cellcolor{gray!10}FMT-L (cfg=2.00) &\cellcolor{gray!10}394M &\cellcolor{gray!10}3.72 &\cellcolor{gray!10}297.54 &\cellcolor{gray!10}0.86 &\cellcolor{gray!10}0.49 \\
         &\cellcolor{gray!10}FMT-XL (cfg=1.75) &\cellcolor{gray!10}893M &\cellcolor{gray!10}2.76 &\cellcolor{gray!10}273.76 &\cellcolor{gray!10}0.84 &\cellcolor{gray!10}0.55 \\
        \midrule
        \multirow{3}{*}{SAR-TS} & \cellcolor{gray!10}FMT-B (cfg=2.00) & \cellcolor{gray!10}125M & \cellcolor{gray!10}7.19 & \cellcolor{gray!10}186.20 & \cellcolor{gray!10}0.85 & \cellcolor{gray!10}0.39 \\
         & \cellcolor{gray!10}FMT-L (cfg=2.00) & \cellcolor{gray!10}394M & \cellcolor{gray!10}4.67 & \cellcolor{gray!10}246.46 & \cellcolor{gray!10}0.84 & \cellcolor{gray!10}0.46 \\
        (random-$16$-random) &\cellcolor{gray!10}FMT-XL (cfg=1.90) &\cellcolor{gray!10}893M &\cellcolor{gray!10}4.01 &\cellcolor{gray!10}250.32 &\cellcolor{gray!10}0.82 &\cellcolor{gray!10}0.50 \\
    \bottomrule
    \end{tabular}
\end{table}
\begin{table}[t]
    \centering
    \begin{minipage}[ht]{0.68\textwidth}
        \scriptsize
        \tabcaption{FID results of training/inference with different order settings. The model is FMT-B.}
        \vspace{2pt}
        \label{tab:order_order}
        \centering
        \renewcommand{\arraystretch}{1.2}
        \addtolength{\tabcolsep}{-2.6pt}
        \begin{tabular}{@{}lcccccccc@{}}
            \toprule
            Training/inference & raster & reversed-raster & roll & reversed-roll & fixed-random & random\\
            \midrule
            raster & \textbf{5.40} & 136.54 & 114.41 & 99.13 & 132.61 & 120.82 \\
            reversed-raster & 133.18 & \textbf{6.01} & 123.47 & 118.67 & 146.48 & 138.29 \\
            roll & 81.93 & 114.23 & \textbf{6.93} & 133.50 & 130.28 & 117.69 \\
            reversed-roll & 125.78 & 134.25 & 155.04 & \textbf{6.44} & 128.62 & 125.56 \\
            fixed-random & 104.24 & 117.23 & 116.58 & 103.03 & \textbf{7.49} & 86.90 \\
            random & 22.95 & 22.91 & 13.66 & 10.32 & 7.83 & \textbf{7.76} \\
            \bottomrule
        \end{tabular}
    \end{minipage}
    \hfill
    \begin{minipage}[ht]{0.3\textwidth}
    \setlength{\abovecaptionskip}{-8pt}
        \centering
        \includegraphics[width=\linewidth]{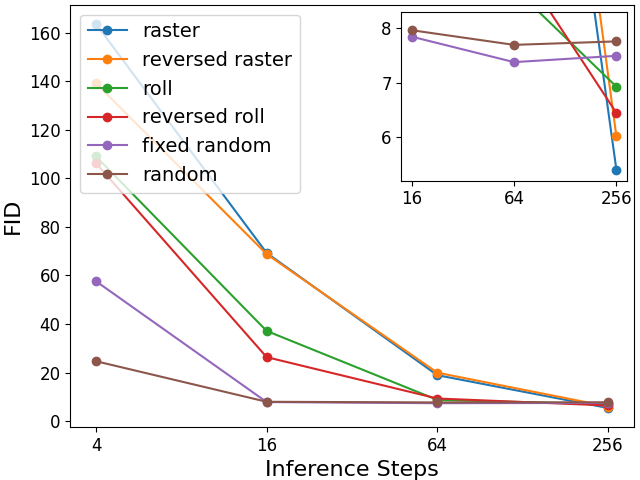}
        \figcaption{Effect of order.}
        \label{fig:order_step}
    \end{minipage}%
\end{table}

Table~\ref{tab:performance} presents a comprehensive comparison of performance across various methods and models, where we train models for each AR setting within the SAR paradigm.

\textbf{SAR as AR.} The raster-$256$-cosine variant of SAR recovers to conventional AR. We evaluate the performance of FMT-B, FMT-L, and FMT-XL, with the results presented in Table~\ref{tab:performance}. Under the same setting (stared in Table~\ref{tab:performance}, directly training at $256\times256$), FMT outperforms LlamaGen under the same model size.

\textbf{SAR as MAR.} SAR recovers to MAR under the `masked modeling' setting. The performance of FMT is also shown in Table~\ref{tab:performance}. 

\textbf{SAR as VAR, analogously.} By customizing the sequence order and output intervals as `next-scale', illustrated on the left side of Fig.~\ref{fig:rearrange}, we derived a rough variant of VAR. The results are presented in Table~\ref{tab:performance}. While this serves primarily as a conceptual example, its performance lags significantly behind that of VAR~\citep{tian2024visual}.

\textbf{Transition states of SAR.} The last three rows of Table~\ref{tab:performance} present the performance ($64$ steps) of a specific design choice in the transition states of SAR, which will be detailed in the ablation study. Compared to FMT under the AR configuration, the performance in this case is somewhat lower. However, models trained under this setting can generalize across inference steps and orders while maintaining their causal features. A straightforward merit is that, we can enable KV cache acceleration while performing few-step inference. A diagram on performance-time trade-off is shown in Fig.~\ref{fig:imagenet_time}, where the inference time is tested by generating a batch of $8$ images on one A100 GPU. We may also apply other causal techniques to promote the performance or efficiency.

\vspace{-1mm}
\subsection{Ablation Study}
\vspace{-1mm}
\textbf{Varying sequence orders in training/inference.} Table~\ref{tab:order_order} presents the results obtained by fixing the output intervals to $1,1,\ldots$ while training and inferring with various sequence orders. It is clear that although position embeddings are used, a fixed sequence order typically does not allow the model to generalize across different inference orders.

\begin{figure}[t]
    \centering
    \begin{minipage}[ht]{0.3\linewidth}
        \centering
        \includegraphics[width=\linewidth]{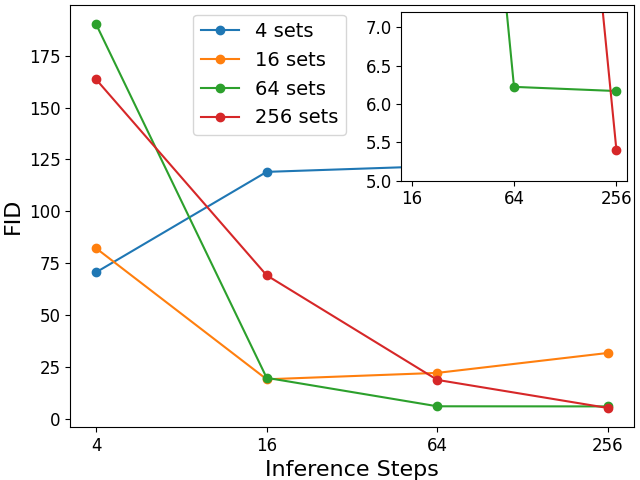}
    \end{minipage}
    \hfill
    \begin{minipage}[ht]{0.3\linewidth}
        \centering
        \includegraphics[width=\linewidth]{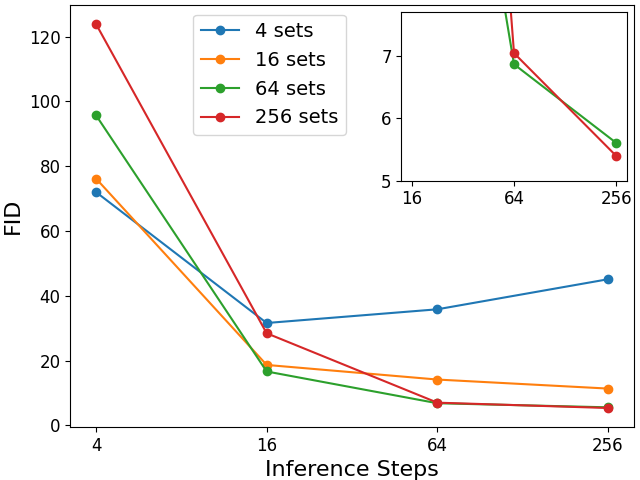}
    \end{minipage}
    \hfill
    \begin{minipage}[ht]{0.3\linewidth}
        \centering
        \includegraphics[width=\linewidth]{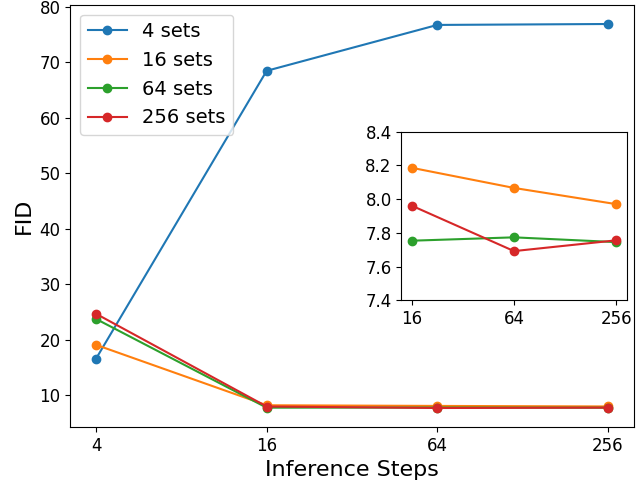}
    \end{minipage}
    \caption{Effect of set numbers when training SAR with (left) raster order and cosine schedule, (middle) raster order and random schedule, and (right) random order and cosine schedule.}
    \label{fig:exp}
\end{figure}

\textbf{Fixed few-step generation.} By fixing the sequence order to the raster order and using a cosine schedule for the intervals, we investigate few-step SAR training by varying only the number of sets. As illustrated on the left of Fig.~\ref{fig:exp}, we observe that, i) since both the order and the schedule are fixed, the best inference performance typically occurs when the number of sets used at inference matches that used in training; ii) from the inset in the upper right, it is evident that only the $64$-set configuration is effective for few-step generation, while the others significantly degrade performance.

\textbf{Randomness in orders enables few-step generalization.} We fix the number of sets at $256$ and the interval schedule to $1,1,\ldots$, varying only the sequence order. As shown in Fig.~\ref{fig:order_step}, models trained with the raster, reversed raster, roll, and reversed roll orders struggle to generalize to few-step generation. In contrast, models trained with a random order demonstrate good generalization across inference steps, albeit at the cost of lower FID scores ($5.40$ FID with raster order vs. $7.76$ FID with random order). It may be surprising that fixing a randomly generated order during training can achieve similar generalization ability to that of a fully random order.

\begin{figure}[t]
    \centering
    \begin{minipage}[ht]{0.3\linewidth}
        \centering
        \includegraphics[width=\linewidth]{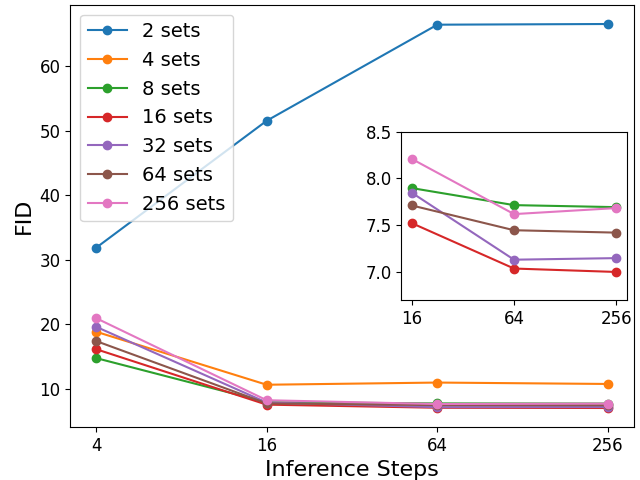}
    \end{minipage}
    \hfill
    \begin{minipage}[ht]{0.3\linewidth}
        \centering
        \includegraphics[width=\linewidth]{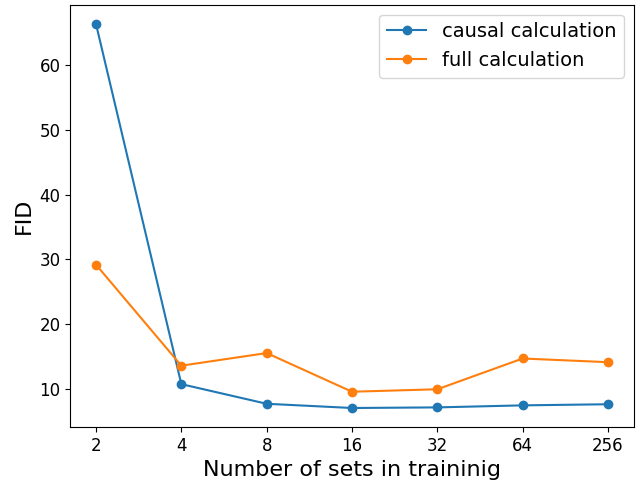}
    \end{minipage}
    \hfill
    \begin{minipage}[ht]{0.31\linewidth}
        \centering
        \includegraphics[width=\linewidth]{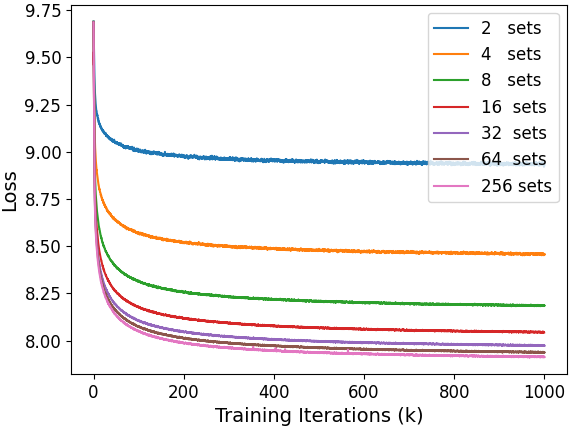}
    \end{minipage}
    \caption{Exploration when sequence order and output schedule are both set as random. Left: performance wrt. number of sets. Middle: after causal training, comparison between causal and full attention calculation. Right: training loss of various set numbers.}
    \label{fig:rxr}
\end{figure}

\textbf{Random output intervals enables few-step generalization.} We fix the sequence order to raster and use a random schedule with varying numbers of sets. The results on the middle of Fig.~\ref{fig:exp} indicate that when the number of sets is large (\textit{e.g.}, $64$ or $256$), random intervals facilitate few-step generalization.

\textbf{The relationship between number of sets and causal learning.} Under the setting of random sequence order, we examine performance in relation to the number of sets. Figures~\ref{fig:exp} (right) and~\ref{fig:rxr} (left) show the results with cosine and random output schedules, respectively. We observe that, with a large number of sets, performance remains stable; however, it declines significantly when the set number decreases to $4$ in the cosine case and $2$ in the random case. Intuitively, to develop a causal model, the model must be trained to predict sets one by one, with more sets indicating a greater degree of causality. If the number of sets is too small, the model struggles to learn causal relationships effectively. Another interesting observation is that, after trained with small number of sets, abandoning causality can help restore performance. As shown in the middle of Fig.~\ref{fig:rxr}, the performance of the model trained with $2$ sets gets better when replacing the causal attention with full attention. However, model trained with other set numbers cannot benefit from full attention, because they receive more sufficient causal learning. The last subfigure of Fig.~\ref{fig:rxr} illustrates the loss curves during training, where the level of loss may be regarded as a measure of training difficulty. The loss of the best-performing configuration, $16$ sets, is situated at a mid-level.

\textbf{Further discussion on the MAR setting of SAR.} There are some details that need to be clarified.
i) In Sec.~\ref{sec:experiments}, we mentioned that the MAR setting is derived based on `random-$2$-random' by only supervising the second set, and using the random strategy in \citet{li2024autoregressive}. From Table~\ref{tab:mar}, Row 1 vs. Row 2 tells us that, with the same model, removing the loss of the first set has little impact on model training; not removing it may even lead to better performance. This fact demonstrates that the transition from $K=2$ to $K=1$ (\textit{i.e.}, MAR) in SAR is smooth. ii) It is worth noting that, in the MAR case the generalized causal masks in the encoder self-attention and decoder cross-attention is equivalent to having none. And only the causal mask in decoder self-attention will affect the training. Intuitively, there is no need to prepare causal mask in training because at inference MAR always conduct global attention. Row 1 vs. Row 3 in Table~\ref{tab:mar} indicates that the existence of causal mask in decoder self-attention hurts the performance. iii) Row 4 is a setting from Fig.~\ref{fig:rxr}. The large discrepancy in performance between Row 2 and Row 4 emphasizes the importance of proper random strategy. This also suggests that our strategy for SAR transition states may not be optimal, which may explain the sub-optimal SAR-TS results in Table~\ref{tab:performance}.

\begin{table}[t]\scriptsize
    \centering
    \begin{minipage}[ht]{0.6\textwidth}
    \tabcaption{Results among detailed MAR settings. The inference process is BERT-like, with full attention.}
    \label{tab:mar}
    \centering
    \renewcommand{\arraystretch}{1.1}
    \addtolength{\tabcolsep}{-2.4pt}
    \begin{tabular}{@{}lccccccc@{}}
    \toprule
        Random Strategy & K & Causal Mask & FID$\downarrow$ & IS$\uparrow$ & Precision$\uparrow$ & Recall$\uparrow$ \\
        \midrule
        MAR~\citep{li2024autoregressive} & 1 & \Checkmark & 8.81 & 148.36 & 0.76 & 0.46 \\
        MAR~\citep{li2024autoregressive} & 2 & \Checkmark & 7.19 & 183.31 & 0.83 & 0.39 \\
        MAR~\citep{li2024autoregressive} & 1 & \XSolidBrush & 6.98 & 222.28 & 0.87 & 0.36 \\
        Equal Probability & 2 & \Checkmark & 29.20 & 46.91 & 0.65 & 0.52 \\
    \bottomrule
    \end{tabular}
    \end{minipage}%
    \hfill
    \begin{minipage}[ht]{0.37\textwidth}
    \setlength{\belowcaptionskip}{2pt}
    \tabcaption{Comparison on inference time with $4096$ tokens and FMT-XL.}
    \label{tab:time}
    \centering
    \renewcommand{\arraystretch}{1.1}
    \addtolength{\tabcolsep}{-4.4pt}
    \begin{tabular}{@{}lcccccccc@{}}
    \toprule
        Setting & KV cache & 64 steps & 128 steps & 4096 steps \\
        \midrule
        AR & \Checkmark & - & - & 174.49s \\
        MAR & \XSolidBrush & 9.66s & 19.22s & 685.77s \\
        SAR-TS & \XSolidBrush & 7.45s & 14.72s & 606.35s \\
        SAR-TS & \Checkmark & 2.82s & 5.78s & 174.49s \\
    \bottomrule
    \end{tabular}
    \end{minipage}
    \vspace{-4mm}
\end{table}
\subsection{Application: Text-to-Image Generation}
\begin{figure}[ht]
    \setlength{\abovecaptionskip}{-6pt}
    \centering
    \includegraphics[width=\linewidth]{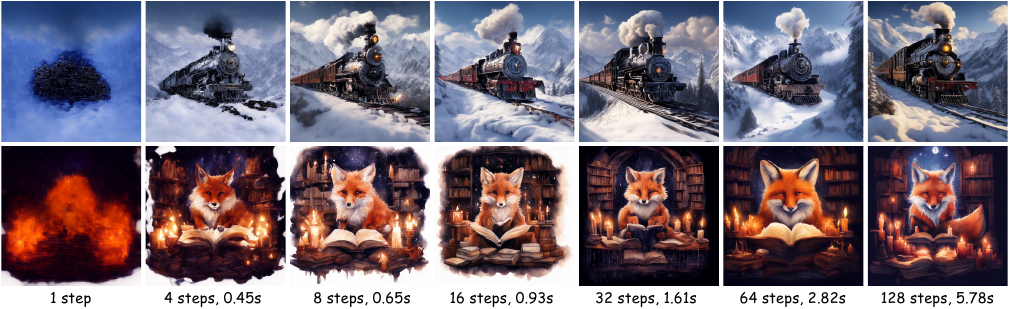}
    \caption{Step number and time cost of Lumina-SAR at $1024\times1024$ (full $4096$ steps cost $174.49s$).}
    \label{fig:t2i_step}
\end{figure}

We leverage the FMT-XL model for text-to-image (T2I) generation. The sequence order and the output schedule are set as random, the best practice with random order in ImageNet experiments. We adopt the training strategy with multiple aspect ratios as in \citet{gao2024lumina,zhuo2024lumina} and the multi-stage policy in \citet{zhuo2024lumina,sun2024autoregressive,chen2024pixart}. Specifically, we set the number of sets as $16$ and the base resolution as $256\times256$ in the first stage, and gradually increase the number of sets and the base resolution by a factor of $2$. The final resolution is $1024$. At each training stage, we group images with different aspect ratios but similar pixel numbers and pad them to the same length. As for the language part, we adopt the Gemma-2B~\citep{team2024gemma} as the text encoder and concatenate the text embedding with the image tokens, with the conventional causal mask like that in Fig.~\ref{fig:comparison} (a1). Other training settings including text-image training data are following \citet{zhuo2024lumina}, and we name our T2I model as Lumina-SAR. As visualized in Fig.~\ref{fig:visualization}, Lumina-SAR can flexibly produce photo-realistic images in arbitrary resolutions.

\textbf{Inference time.} We examine the time cost of Lumina-SAR for generating one image using one A100 GPU, as illustrated in Fig.~\ref{fig:t2i_step}. We observe that Lumina-SAR begins to produce meaningful images at around $4$ to $8$ steps. With $64$ to $128$ steps, it can deliver high-quality outputs, requiring a processing time of only $3$ to $6$ seconds. Typically, the full $4096$ steps take $>60$ times longer than that required for $64$ steps. A detailed comparison of inference times among AR, MAR, and SAR-TS models is presented in Table~\ref{tab:time}. To ensure a fair comparison, we consistently use FMT-XL with a resolution of $1024 \times 1024$, varying only the inference manner. Notably, in the transformer decoder, MAR applies global attention across all tokens, while the number of tokens processed in AR and SAR-TS increases gradually. Consequently, even with KV cache disabled, the inference time for SAR-TS is shorter than that of MAR; when KV cache is enabled, SAR-TS is three times faster than MAR with $64$ or $128$ steps.
\begin{figure}[ht]
    \setlength{\abovecaptionskip}{-6pt}
    \centering
    \includegraphics[width=\linewidth]{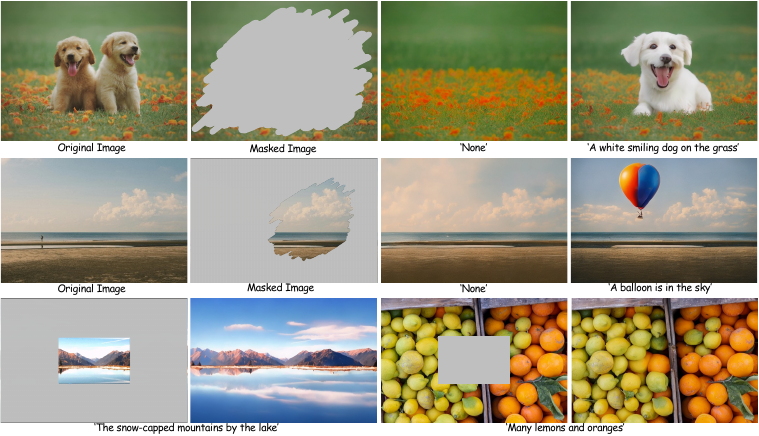}
    \caption{Zero-shot image painting with Lumina-SAR. Gray color indicates masked regions, and the text enclosed in quotes represents the input text prompt.}
    \label{fig:painting}
\end{figure}

\textbf{Zero-shot image painting.} One of the advantages of using random sequence orders is the flexibility in inference order, which facilitates image editing tasks such as image inpainting and outpainting. This is an important feature that AR lacks but MAR~\citep{chang2022maskgit} includes. To validate the painting ability of SAR-TS models, we perform zero-shot painting with Lumina-SAR. Several instances is shown in Fig.~\ref{fig:painting}, where the mask can be any shape.

\section{Conclusion}
\label{sec:conclusion}

In this work, we propose Set AutoRegressive Modeling (SAR), a new AR paradigm with a broader deign space to freely customize the AR training and inference processes. SAR incorporates existing AR variants with flexible sequence order and output intervals. For SAR, we also develop a preliminary architecture called the Fully Masked Transformer. We carefully explore the properties of SAR, with a particular focus on the intermediate states, which integrates advantages of both AR and MAR models. To further validate the generation potential at the transition states, we train a text-to-image model capable of generating high-quality diverse images.

\textbf{Limitation and future work.} As a newly emerging paradigm, the exploration of SAR in this paper is limited, particularly concerning the performance of intermediate states on ImageNet. Future work may focus on developing better training and inference schedules, designing model architectures more compatible with SAR, and exploring the application of SAR across additional modalities.

\bibliography{iclr2025_conference}
\bibliographystyle{iclr2025_conference}

\newpage
\appendix
\section{Appendix}

\subsection{More Visualizations of Class-conditioned Generation on ImageNet} 
\begin{figure}[ht]
    \setlength{\abovecaptionskip}{-6pt}
    \centering
    \includegraphics[width=\linewidth]{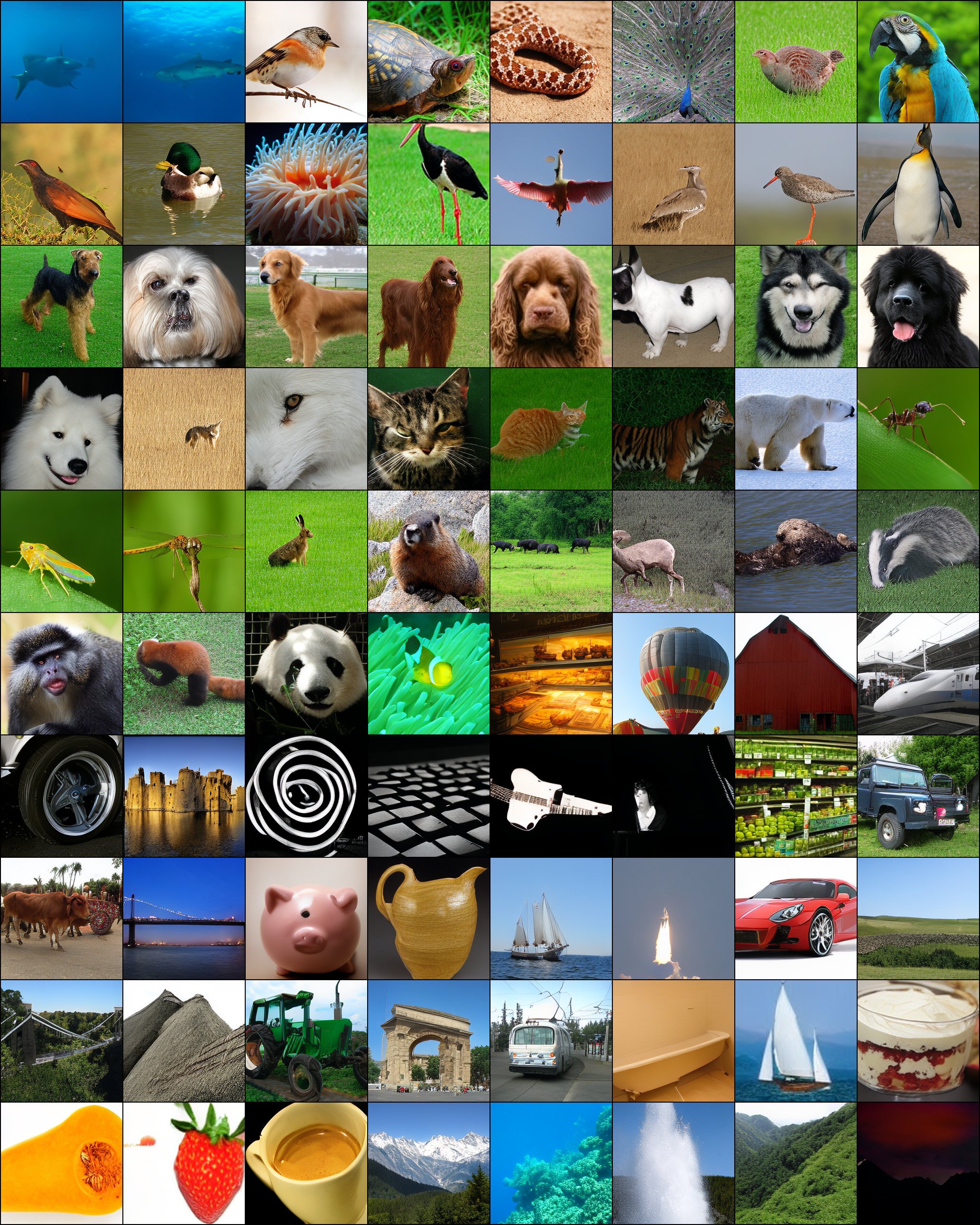}
    \caption{Samples generated by FMT-XL trained with SAR, random-$16$-random.}
    \label{fig:r16r_xl}
\end{figure}

\begin{figure}[ht]
    \setlength{\abovecaptionskip}{-6pt}
    \centering
    \includegraphics[width=\linewidth]{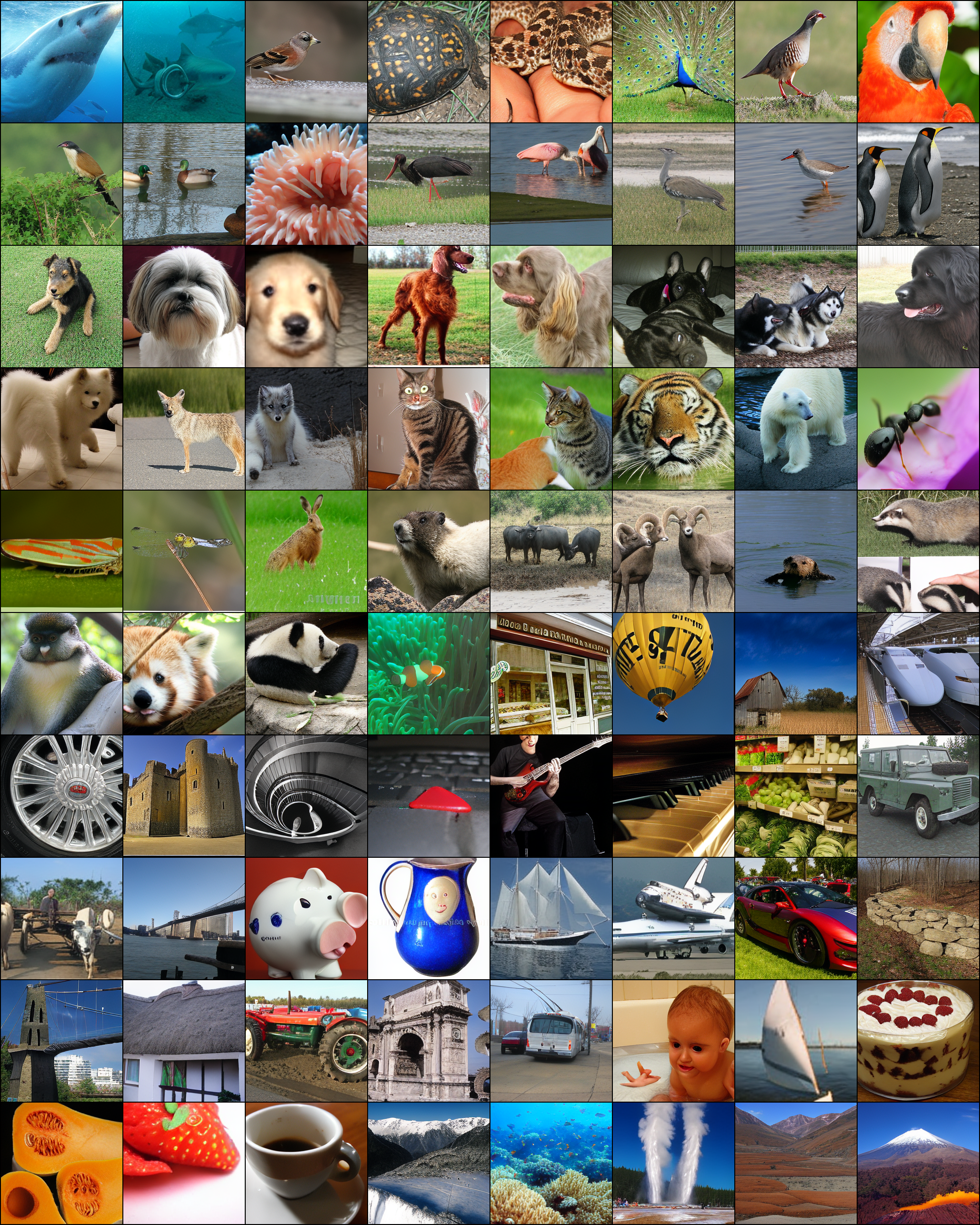}
    \caption{Samples generated by FMT-XL trained with SAR, raster-$256$-cosine (\textit{i.e.}, classical AR).}
    \label{fig:ntp_xl}
\end{figure}

For $256\times 256$ image generation on ImageNet, we generate some random samples that are not cherry picked. Fig.~\ref{fig:r16r_xl} and Fig.~\ref{fig:ntp_xl} exhibit samples produced by FMT-XL trained under random-$16$-random and raster-$256$-cosine settings, respectively.
\begin{figure}[ht]
    \setlength{\abovecaptionskip}{-6pt}
    \centering
    \includegraphics[width=\linewidth]{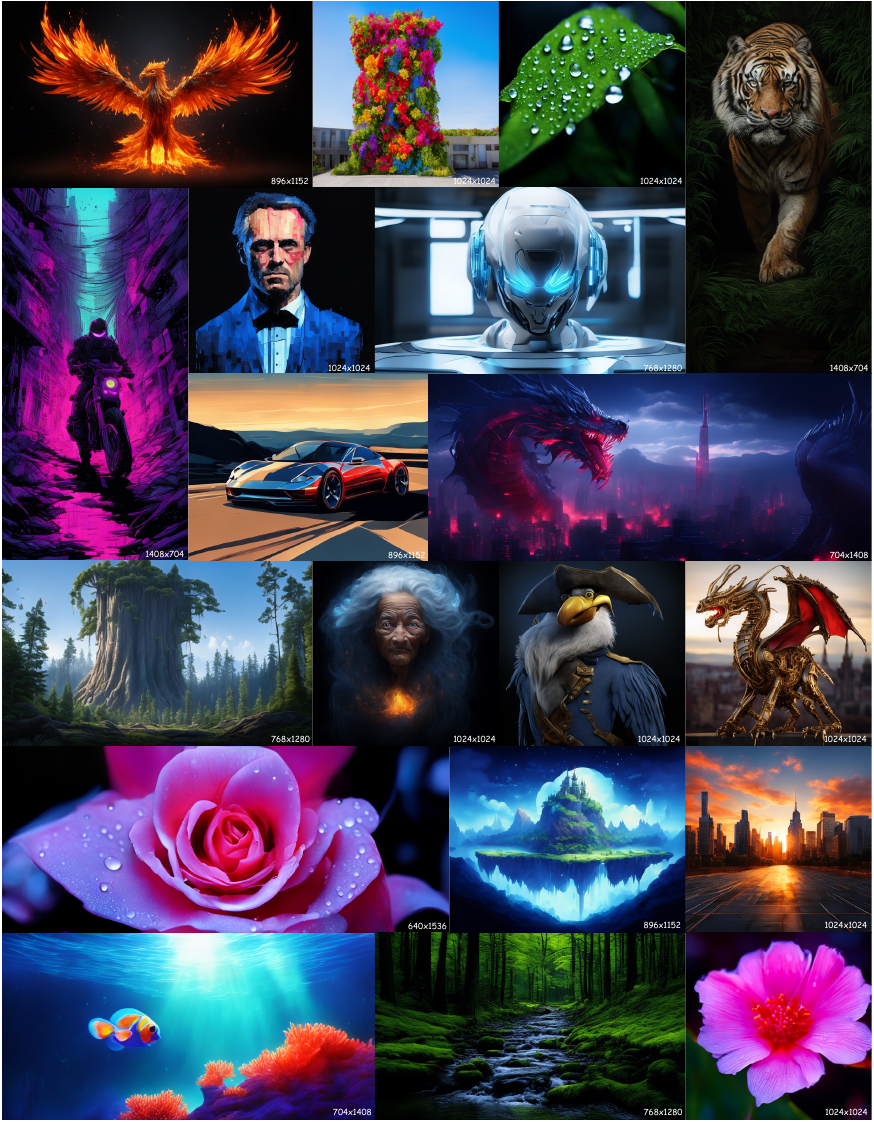}
    \caption{Samples generated by Lumina-SAR. The model is FMT-XL trained under the random-$x$-random setting of SAR, where $x$ is set as $16$, $32$ and $64$ at the stage of $256\times256$, $512\times512$ and $1024\times1024$ respectively.}
    \label{fig:add_t2i}
\end{figure}

\subsection{More Visualizations on T2I Image Synthesis}
We provide additional visualizations generated by Lumina-SAR, and show them in Fig.~\ref{fig:add_t2i}. The number of inference steps is $64$.
\begin{figure}[ht]
    \setlength{\abovecaptionskip}{-6pt}
    \centering
    \includegraphics[width=\linewidth]{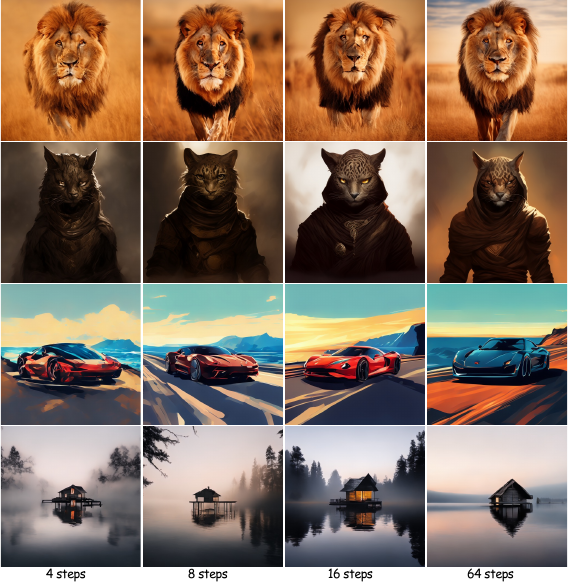}
    \caption{Samples generated by Lumina-SAR with $4$, $8$, $16$ and $64$ inference steps.}
    \label{fig:add_t2i_step}
\end{figure}
\subsection{More Instances on Few-step Text-to-image Generation}
We provide more T2I examples when sampling with $4$, $8$, $16$, and $64$ steps. As shown in Fig.~\ref{fig:add_t2i_step}, the generation quality drops slightly with $16$ steps, and becomes much worse with $4$ or $8$ steps. Hence, we recommend a step number of $64$ for high-quality outputs.
\subsection{Details on Fully Masked Transformer}
With fixed resolution, the position embedding as input to the decoder can be either learnable or fixed, such as sine embedding. The performances are similar between learned and sine position embeddings in class-conditioned generation. In the T2I model, we use sine embedding to accommodate training with multiply aspect ratios: after each input image is fed into FMT, we first generate its sine embedding.
Similar to LlamaGen~\citep{sun2024autoregressive}, we use RoPE~\citealp{su2024roformer} to enable the position-aware interaction. Both the position embedding and RoPE are rearranged like what is done to the input tokens according to the sequence order, such that the positions are aligned.

\textbf{Relation between Fully Masked Transformer and the transformer in \citet{vaswani2017attention}.} Structurally, there are two more generalized causal masks in FMT (at encoder self-attention and decoder cross-attention) to than in \citet{vaswani2017attention}. Functionally, the encoder in vanilla transformer is used to encode the context (\textit{e.g.}, the question in question-and-answer tasks, and the class/text tokens in our case), and the decoder serves as the token generator. In FMT, the encoder and decoder together serve as the token generator, while the encoder also functions as encoding seen tokens (including the context and generated tokens). When regarding the transformer as a black box, FMT can work as a classical decoder-only transformer like Llama~\citep{touvron2023llama}.

\subsection{On Data Augmentation of SAR-TS Training on ImageNet}
\begin{table}[!t] \scriptsize
    \caption{The effect of random crop augmentation in training for the SAR random-$16$-random setting. The models are trained for $300$ epochs, and the number of inference steps is $64$.}
    \label{tab:randomcrop}
    \centering
    \renewcommand{\arraystretch}{1.1}
    \addtolength{\tabcolsep}{8.6pt}
    \begin{tabular}{@{}lcccccccc@{}}
    \toprule
        Model & \#Params & Random Crop & FID$\downarrow$ & IS$\uparrow$ & Precision$\uparrow$ & Recall$\uparrow$ \\
        \midrule
         FMT-B (cfg=2.00) & 125M & \Checkmark & 7.04 & 182.01 & 0.84 & 0.40 \\
         FMT-B (cfg=2.00) & 125M & \XSolidBrush & 7.19 & 186.20 & 0.85 & 0.39 \\
         FMT-L (cfg=2.00) & 394M & \Checkmark & 4.75 & 261.27 & 0.84 & 0.46 \\
         FMT-L (cfg=2.00) & 394M & \XSolidBrush & 4.67 & 246.46 & 0.84 & 0.46 \\
         FMT-XL (cfg=1.90) & 893M & \Checkmark & 4.24 & 249.23 & 0.82 & 0.51 \\
         FMT-XL (cfg=1.90) & 893M & \XSolidBrush & 4.01 & 250.32 & 0.82 & 0.50 \\
    \bottomrule
    \end{tabular}
\end{table}
\begin{figure}[ht]
    \setlength{\abovecaptionskip}{-6pt}
    \centering
    \includegraphics[width=\linewidth]{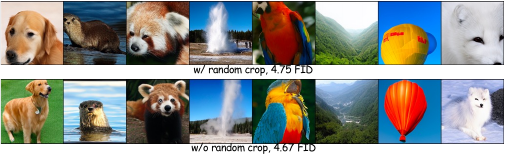}
    \caption{The framing misalignment issue caused by random crop augmentation when training random-$16$-random models.}
    \label{fig:randomcrop}
\end{figure}
In all experiments except SAR-TS models in Table~\ref{tab:performance}, we adopted random crop augmentation following~\citep{sun2024autoregressive}. For SAR-TS models, we find that they are sensitive to random crop augmentation, frequently encountering framing misalignment issues in image generation. Some randomly generated examples by FMT-L are shown in Fig.~\ref{fig:randomcrop}. In a batch of eight simultaneously generated images, the first, third, fifth, seventh, and eighth images exhibit this misalignment issue. Our experiments indicate that while random crop augmentation is effective for smaller models (FMT-B), it negatively impacts the FID scores of larger models (FMT-L, FMT-XL), as shown in Table~\ref{tab:randomcrop}. By comparing generated images in Fig.~\ref{fig:randomcrop}, we observe that random crop augmentation contributes to the framing misalignment problem; removing it mitigates this issue and improves the FID score (but to some extent hurts the visual quality as perceived by human eyes). And as a result, we report the quantitative results of SAR-TS models without random crop in Table~\ref{tab:performance}.

\subsection{The Effect of Evaluation Configurations} 
\begin{figure}[ht]
    \centering
    \begin{minipage}[ht]{0.48\linewidth}
        \centering
        \includegraphics[width=\linewidth]{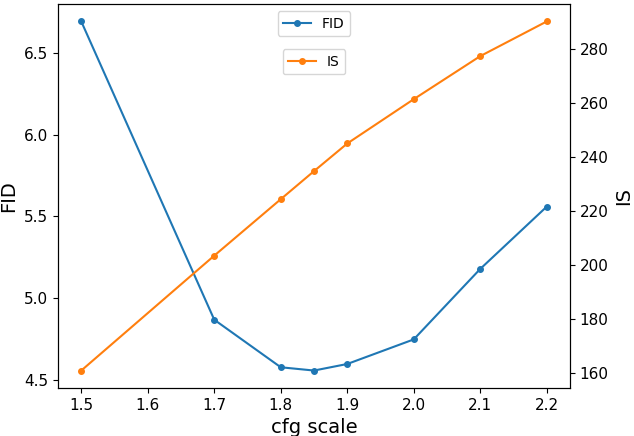}
    \end{minipage}
    \hfill
    \begin{minipage}[ht]{0.48\linewidth}
        \centering
        \includegraphics[width=\linewidth]{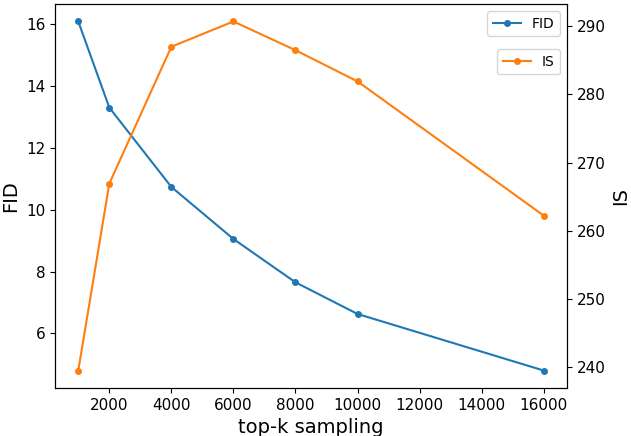}
    \end{minipage}
    \caption{The effect of cfg scale (left), and top-k sampling (right).}
    \label{fig:sample_config}
\end{figure}

We provide the results when adjusting the scale of classifier-free guidance and the top-k values in Fig.~\ref{fig:sample_config}, where we use FMT-L trained under the random-$16$-random setting for $300$ epochs and the number of sampling steps is set to $64$. The inference behavior is similar to that of classical AR models.

\end{document}